\def\year{2017}\relax
\documentclass[letterpaper]{article}
\usepackage{aaai17}
\usepackage{times}
\usepackage{helvet}
\usepackage{courier}
\usepackage{graphicx}
\usepackage{CJK}
\usepackage{multirow}
\usepackage{amsmath}
\begin{document}
 \begin{CJK*}{UTF8}{gbsn}
% The file aaai.sty is the style file for AAAI Press
% proceedings, working notes, and technical reports.
%
\title{Structure Regularized Bidirectional Recurrent Convolutional Neural Network for Relation Classification }

%\author{Ji Wen, Xu Sun, Xuancheng Ren \\ MOE Key Laboratory of Computational Linguistics, %Peking University \\ School of Electronics Engineering and Computer Science, Peking %University\\      {wenjics, xusun, renxc}@pku.edu.cn }

\author{Ji Wen \\ School of Electronics Engineering and Computer Science, Peking University\\      {wenjics}@pku.edu.cn }

\maketitle
\begin{abstract}
Relation classification is an important semantic processing task in the field of natural language processing (NLP). In this paper, we present a novel model, Structure Regularized Bidirectional Recurrent Convolutional Neural Network(SR-BRCNN), to classify the relation of two entities in a sentence, and the new dataset of Chinese Sanwen for named entity recognition and relation classification. Some state-of-the-art systems concentrate on modeling the shortest dependency path (SDP) between two entities leveraging convolutional or recurrent neural networks. We further explore how to make full use of the dependency relations information in the SDP and how to improve the model by the method of structure regularization. We propose a structure regularized model to learn relation representations along the SDP extracted from the forest formed by the structure regularized dependency tree, which benefits reducing the complexity of the whole model and helps improve the $F_{1}$ score by 10.3. Experimental results show that our method outperforms the state-of-the-art approaches on the Chinese Sanwen task and performs as well on the SemEval-2010 Task 8 dataset\footnote{The Chinese Sanwen corpus this paper developed and used will be released in the further. }.

\end{abstract}

\section{Introduction}

Relation classification aims to classify the semantic relations between two entities in a sentence. For instance, in the sentence ``The [burst]e1 been caused by water hammer [pressure]e2 has'', entities ``burst'' and ``pressure'' are of relation Cause-Effect(e2, e1). Relation classification plays a key role in robust knowledge extraction, and has become a hot research topic in recent years. 

Recently, more attention has been paid to modeling the shortest dependency path (SDP) of sentences. \cite{liu2015dependency} developed a dependency-based neural network, in which a convolutional neural network has been used to capture features on the shortest path and a recursive neural network is designed to model subtrees. \cite{xu2015classifying} applied long short term memory (LSTM) based recurrent neural networks (RNNs) along the shortest dependency path.

\subsection{Relation Classification}
Relation classification is the task of identifying the semantic relation holding between two nominal entities in text. It is a crucial component in natural language processing systems that need to mine explicit facts from text, e.g. for various information extraction applications as well as for question answering and knowledge base completion.

Word segmentation and named entity recognition are usually applied to the sentences, on which relation classification will be conducted. Since this task has crucial connections to NER and word segmentation information, improving the accuracy on NER and word segmentation is helpful to this task. For Chinese word segmentation, \citeauthor{conf/acl/SunWL12} (\citeyear{conf/acl/SunWL12}) presented a joint model for Chinese word segmentation and new word detection. \citeauthor{SunZMTT13} (\citeyear{Xu2016Dependency}) focus on Chinese word segmentation by systematically incorporating non-local information based on latent variables and word-level features. Methods using neural networks are raised for word segmentation. \citeauthor{Xu2016Dependency} (\citeyear{Xu2016Dependency}) proposed a dependency-based gated recursive neural network to integrate local features with long distance dependencies for better performance. \citeauthor{XuS17} (\citeyear{XuS17})  propose a transfer learning method to improve low-resource word segmentation by leveraging high-resource corpora.
After word segmentation, named entity recognition can help recognize named entities and classify these entities for further experiments. \citeauthor{DBLP:journals/corr/HeS16} (\citeyear{DBLP:journals/corr/HeS16}) and \citeauthor{DBLP:conf/aaai/HeS17} (\citeyear{DBLP:conf/aaai/HeS17}) proposed novel method named entity recognition for Chinese social media.

It is not surprising that numerous features and kernel-based approaches have been proposed,
many of which rely on a full-fledged NLP stack, including morphological analysis, POS tagging, dependency parsing, and occasionally semantic analysis, as well as on knowledge resources to capture lexical and semantic features. Bunescu and Mooney\shortcite{bunescu2005shortest} first used shortest dependency paths between two entities to capture the predicate-argument sequences, which provided strong evidence for relation classification.In recent years, we have seen a move towards deep architectures that are capable of learning relevant representations and features without extensive manual feature engineering or use of external resources. A number of convolutional neural network (CNN), recurrent neural network (RNN), and other neural architectures have been proposed for relation classification. Still, these models often fail to identify critical cues, and many of them still require an external dependency parser.

\subsection{Chinese Sanwen}
Most English essays are organized deliberately according to the Westerners' analytic and logic way of thinking. Usually, there are internal logic relations in a sentence and paragraph.  However, Chinese Sanwen tend to express intuition and feelings rather than conducting analysis and argumentation. The following parts respectively state the problems in processing Chinese Sanwen.

First, Chinese Sanwen has a wide range of topics. There are always a large number of multiple entities in Sanwen. Existing entity dictionary could only cover a small set of entities in Sanwen. Little is done to help recognize entities in Sanwen with additional database. Besides the variety and numbers of entities, many Sanwen express feelings in a subtle way, thus they are inclined to employ fuzzy words, making it more difficult to recognize entities.

Second, Chinese Sanwen usually are not organized very logically, no matter among paragraphs or sentences. Sanwen tend to use various forms of sentence to create free and nimble feelings. The implicit expression will make readers chew the words and ponder feelings. 

Third, sentence structure of Chinese Sanwen is very flexible. Unlike the multifarious and unfixed sentence structure of Chinese Sanwen the sentences of English essays are comparatively formal and fixed in that most sentences contain a subject which is noticeable and a verb which is completely subject to the subject. Besides, plenty of connectives implying logical relationships may be employed to make sentences more compact and relations of sentences more distinct. On the contrary, the sentences in Chinese Sanwen” are not associated with each other by evident conjunctions, Besides, Chinese is a topic-prominent language, subject is usually covert and the verb is relatively flexible in form.\\

In this paper, we focus on the study of applying structure regularization to the relation classification task.  To summarize, the contributions of this paper are as followings:
\begin{itemize}
	\item To the best of our knowledge, we are the first to develop a corpus of Chinese Sanwen. The corpus contains 726 articles. The whole size of the corpus is 13.2 MB. It helps alleviate the dilemma of lack of corpus in Chinese Named Entity Recognition and Relation Classification.
	\item We develop tree-based structure regularization and make a progress on the task of relation classification.  The method of structure regularization is normally used on the structure of sequences, while we find a way to realize it on the structure of trees. Comparing to the origin model, our model improves the $F_{1}$ score by 10.3.
	
\end{itemize}

\begin{figure}
	\centering
	\centerline{\includegraphics[width= 9cm]{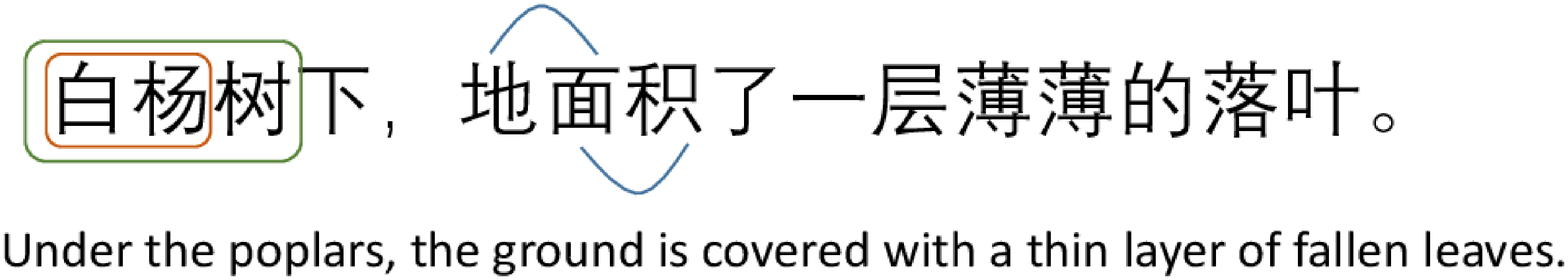}}
	\caption{An example of the difficulty to process Chinese sanwen}\label{fig1}
\end{figure}

\section{Background}

\subsection{Long Short Term Memory Network}

Recently, long short term memory network (LSTM)(\cite{hochreiter1997long}) has widely applied to many tasks, like Chinese word segmentation, POS tagging, neural machine translation and so on. In this work, we use the bi-directional long short term memory network (Bi-LSTM) to capture long and short information which is essential for CWS. The LSTM recurrent cell is controlled by three 'gates', namely input gate ${i^{(t)}}$, forget gate $f^{(t)}$ and output gate $o^{(t)}$.  The inputs of LSTM cell are $x^{(t)}$, $s^{(t-1)}$ and $h^{(t-1)}$.
\begin{eqnarray}
g^{(t)}=\tanh(w_{gx} x^{(t)}+w_{gh} h^{(t-1)}+b_{g})
\end{eqnarray}
\begin{eqnarray}
i^{(t)}=sigmoid(w_{ix} x^{(t)}+w_{ih} h^{(t-1)}+b_{i})
\end{eqnarray}
\begin{eqnarray}
f^{(t)}=sigmoid(w_{fx} x^{(t)}+w_{fh} h^{(t-1)}+b_f )
\end{eqnarray}
\begin{eqnarray}
o^{(t)}=sigmoid(w_{ox} x^{(t)}+w_{oh} h^{(t-1)}+b_o)
\end{eqnarray}
The core of LSTM cell is $s^{(t)}$, which is computed by the former state $s^{(t-1)}$ and two gates, ${i^{(t)}}$ and $f^{(t)}$.
\begin{eqnarray}
s^{(t)}=g^{(t)}\odot i^{(t)}+s^{(t-1)}\odot f^{(t)}
\end{eqnarray}
 The output of LSTM cell $h^{(t)}$ is calculated by $s^{(t)}$ and $o^{(t)}$.
\begin{eqnarray}
h^{(t)}=\tanh(s^{(t)}\odot o^{(t})
\end{eqnarray}

\subsection{Structure Regularization}
Structured prediction models are popularly used to solve structure dependent problems in a wide variety of application domains including natural language processing, bioinformatics, speech recognition, and computer vision. Recently, many existing systems on structured prediction focus on increasing the level of structural dependencies within the model. It is argued that this trend could have been misdirected, because the study suggests that complex structures are actually harmful to model accuracy from the work of \cite{sun2014structure}. While it is obvious that intensive structural dependencies can effectively incorporate structural information, it is less obvious that intensive structural dependencies have a drawback of increasing the generalization risk, because more complex structures are easier to suffer from overfitting. Since this type of overfitting is caused by structure complexity, it can hardly be solved by ordinary regularization methods such as L2 and L1 regularization schemes, which is only for controlling weight complexity.

\section{Chinese Sanwen Corpus}

\begin{figure}
	\centering
	\includegraphics[width= 8.5cm]{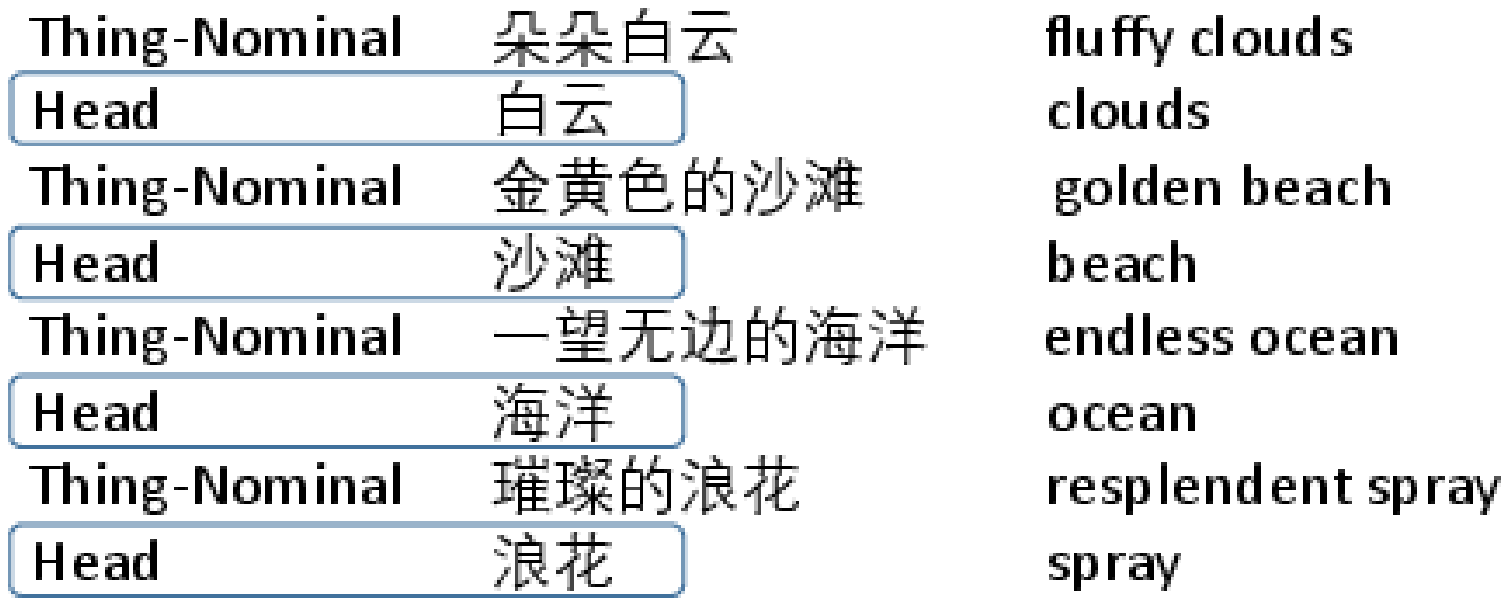}
	\caption{Long entities in Chinese Sanwen}\label{fig2}
\end{figure}

To build the Chinese Sanwen dataset is very difficult. Due to the ambiguity of entities and the difficulty to process Sanwen. Nouns in Sanwen usually have various modifier. Although it improves the beauty of literature, it brings much difficulty to many NLP tasks. Some entities remains a obscure bound to whether contain the modifier of it.  And the decision to whether contain it may cause a great difference of final result. For example, there are different opinions to label the entity in the sentence  shown in Figure~\ref{fig1}. Some people consider``树''(trees) as an entity. However, ``白杨树''(polars) is exactly a specific species. It is also reasonable to treat ``白杨树''(polars) as an entity. Situations like this are very common in Chinese Sanwen dataset. They brings much difficulty to construct a reliable corpus.

Besides the ambiguity of entities, Sanwen itself is difficult to deal with. Many basic Natural Language Processing tasks can not obtain a satisfied performance on Sanwen compared to other corpora. Unlike English, Chinese does not have boundary of words. A basic task for processing Chinese is segment. Still take the sentence as an example, overlapping ambiguity cause the main difficulty to segment the sentence. ``地面''(ground) and ``面积''(area) are both legal words in Chinese. However, the result of segment usually has a major impact to the following tasks. In relation classification, we must recognize the entity first. To recognize a entity, we must determine its boundary. The segmentation ambiguity causes much difficulty to recognize entities.

\begin{figure*}
	\centering
	\includegraphics[width= 16.5cm]{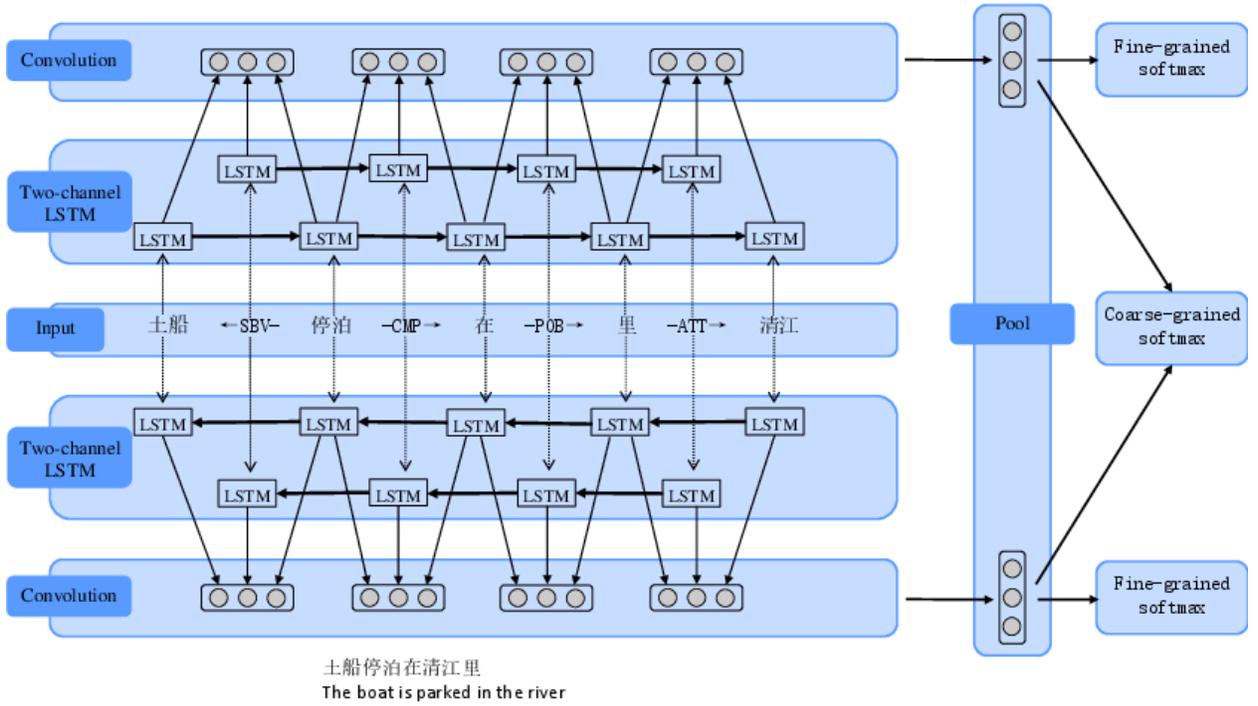}
	\caption{The overall architecture of BRCNN. Two-Channel recurrent neural networks with LSTM units pick up information along the shortest dependency path, and inversely at the same time. Convolution layers are applied to extract local features from the dependency units. In the example, we conduct the process between the entities of ``土船''(boat) and ``清江''(river)}\label{fig3}
\end{figure*}

We build the corpora through a few steps. First, we unify the standard to label entities. In the first edition, we try to label all words that may be treated as entities. To do this, we will find more potential problems when to determine an entity. The rest work will then make a strict screen. Second, exchange the different parts of corpus and examine it carefully. Although we have unify the standard to label entities, different people may have different understandings of whether a word should be treated as an entity. So we exchange the labelled part of corpus to find the labelling difference. Then We correct the mistaken labels and remark the ambiguous labels. Besides, we find there are many long entities in corpus. Some examples are shown in Figure~\ref{fig2}. Our model are base on neural networks, long distance will weaken the information in networks. These entities contains too many words are simplified. We remove the modifier of them and only keep the key words ``head''. Cause long entities are harmful to recognize and they will effect the following work.  

At this point, we have the embryonic form of Sanwen corpus. Finally, we use a program to label the corpus automatically. Entities in the part, which we have labelled and examined carefully, are usually reliable. These entities form an entity dictionary. We then try to match the entities in every article with the entity dictionary to prevent some entities have been omitted. 

Our final corpus contain 726 articles. In prior work, Chinese Sanwen corpus is very rare. Due to the difficulty to process them. Many NLP tasks can not obtain a satisfied result on Sanwen compared to other corpus. Chinese Sanwen is always loosely organized and has various sentence structure. It remains a difficult field in natural language processing. However, Sanwen is a important part of Chinese literature. Understanding Sanwen is of great significance to research Chinese literature.

\section{Structure Regularization BRCNN}

\subsection{Basic BRCNN}

The Bidirectional Recurrent Convolutional Neural Network(BCRNN) model is used to learn representations with bidirectional information along the shortest dependency path(SDP) forwards and backwards at the same time. 

Given a sentence and its dependency tree, we build our neural network on its SDP extracted from tree. Along the SDP, recurrent neural networks with long short term memory units are applied to learn hidden representations of words and dependency relations respectively. A convolution layer is applied to capture local features from hidden representations of every two neighbor words and the dependency relations between them. A max pooling layer thereafter gathers information from local features of the SDP and the inverse SDP. We have a softmax output layer after pooling layer for classification in the unidirectional model RCNN.

On the basis of RCNN model, we build a bidirectional architecture BRCNN taking the SDP and the inverse SDP of a sentence as input. During the training stage of a (K+1)-relation task, two fine-grained softmax classifiers of RCNNs do a (2K + 1)-class classification respectively. The pooling layers of two RCNNs are concatenated and a coarse-grained softmax output layer is followed to do a (K + 1)-class classification. The final (2K+1)-class distribution is the combination of two (2K+1)-class distributions provided by fine grained classifiers respectively during the testing stage.

Bunescu and Mooney (2005) first used shortest dependency paths between two entities to capture the predicate-argument sequences, which provided strong evidence for relation classification. Each two neighbour words are linked by a dependency relation in shortest dependency path. The order of the words will affect the meaning of relations. Single direction of relation may not reflect all information in context. Thus, we employ a bidirectional recurrent convolutional neural network to capture more information from the sentence. The corresponding relation keeps the same when we inverse the shortest dependency path.

Due to the limitations of recurrent neural networks to capture longterm dependencies, we employ LSTM in our work. LSTM performs better in tasks where long dependencies are needed. Some gating units are designed in a LSTM cell. Each of them are in charge of specific functions. We use two bidirectional LSTMs to capture the features of words and relations separately. Word embedding and relation embedding are initialized with two look up tables. Then we can get a real-valued vector of every word and relation according to their index. Embeddings of words are pre-trained on Gigaword with word2vec. In recurrent neural networks, the input is the current embedding $x_t$ and it previous state $h_{t-1}$. For the LSTM that captures word information, $x_t$ is the word embedding and for the LSTM that captures relation information, $x_t$ is the relation embedding. The current step output is denoted as $h_t$. We consider it a representation of all information until this time step. More common, a bidirectional LSTM is used to capture the information previous and later information, as we did in this work.

Convolutional neural network performs well in capturing local features. After we obtain presentations of words and relations, we concatenate them to get a presentation of a complete dependency unit. The hidden state of a relation is denoted as $r_{ab}$. Words on its sides have the hidden states denoted as $h_a$ and $h_b$.  [$h_a$ $h_{ab}$ $h_b$] denotes the presentation of a dependency unit $L_{ab}$. Then we utilize a convolution layer upon the concatenation. We have
\begin{equation}
L_{ab} = f(W_{con} \cdot [h_a \oplus h'_{ab} \oplus h_b] + b_{con})
\end{equation}

where $W_{con}$ is the weight matrix and $b_{con}$ is a bias term. We choose $tanh$ as our activation function and a max pooling followed.

Recurrent neural networks have a long memory, while it causes a distance bias problem. Where inputs are exactly the same may have different representations due to the position in a sentence. However, entities and key components could appear anywhere in a SDP. Thus, two RCNNs pick up information along the SDP and it reverse. A coarse-grained softmax classifier is applied on the global representations $\overrightarrow{G}$ and $ \overleftarrow{G}$. Two find-grained softmax classifier are applied to to give a more detailed prediction of (2K+1) class.
\begin{equation}
\overrightarrow{y} = softmax(W_{f} \cdot \overrightarrow{G} + b_{f}) 
\end{equation}
\begin{equation}
\overleftarrow{y} = softmax(W_{f} \cdot \overleftarrow{G} + b_{f})
\end{equation}

During training, our objective is the penalized cross-entropy of three classifiers. Formally,
\begin{equation}
\begin{split}
J = &\sum_{i=1}^{2K+1}\overrightarrow{t_{i}}log\overrightarrow{y_{i}}+\sum_{i=1}^{2K+1}\overleftarrow{t_{i}}log\overleftarrow{y_{i}} \\
&+\sum_{i=1}^{K}{t_{i}}logy_{i}+\lambda \cdot \Arrowvert \theta \Arrowvert^{2}
\end{split}
\end{equation}
When decoding, the final prediction is a combination of $\overrightarrow{y}$ and  $\overleftarrow{y}$
\begin{equation}
y_{test} = \alpha \cdot\overrightarrow{y} + (1 - \alpha) \cdot z(\overleftarrow{y})
\end{equation}

\subsection{Structure Regularized BRCNN}
The BRCNN model can handle the relation classification task well, but it still remains some weakness, especially dealing with long sentences with complicated structures. The longer the sentences are, the longer the SDPs are.The more complicated the dependency trees are, the more irrelevant words will the SDPs consist. In order to get better SDPs, we propose the Structure Regularized BRCNN.

The idea of structure regularization is straightforward, which is making the structure less complicated so that the model will be free from overfitting or noises of the training and testing data. Unlike the regularization on parameters modifying the loss function, the structure regularization is realized by modifying the structure of the model itself. For example, structure regularization on sequence labelling tasks, usually cuts the sequence into random length first, then the model builds on these pieces of sequences.

As for structure regularized BRCNN, we conduct structure regularization on the dependency tree of the sentences. Based on the certain rule, several nodes in the dependency tree are selected. The subtrees under these selected nodes are cut from the whole dependency tree.With these selected nodes as the roots, these cut subtrees form a forest as well as the original dependency tree, which is much smaller than it was. The forest will be connected by lining the roots of the trees of the forest. Traditional SDP is extracted directly from the dependency tree, while in our model, the SDP is extracted from the lined forest. We call these kind of SDPs as SR-SDP. Finally, we build our BRCNN model on the SR-SDP. 

\begin{table*}[!hbt]
\centering
\begin{tabular}{c|c|c|c}
\hline
\multicolumn{1}{c|}{\multirow{1}{*}{}}&\multicolumn{1}{c|}{\multirow{1}{*}{Models}}&Information&\multicolumn{1}{c}{\multirow{1}{*}{ $F_{1}$ }}\\

\hline
\multicolumn{1}{c|}{\multirow{4}{*}{\textbf{Baselines}}}

&SVM&Word embeddings, NER, WordNet, HowNet,  &48.9\\
&\cite{Hendrickx2010}&POS, dependency parse, Google n-gram&\\

\cline{2-4}

&RNN&Word embeddings&48.3\\
&\cite{socher2011semi}&+ POS, NER, WordNet&49.1\\

\cline{2-4}

&CNN&Word embeddings&47.6\\
&\cite{zeng2014relation}&+ word position embeddings, NER, WordNet&52.4\\

\cline{2-4}

&CR-CNN&Word embeddings&52.7\\
&\cite{santos2015classifying}&+ word position embeddings&54.1\\

\cline{2-4}

&SDP-LSTM&Word embeddings&54.9\\
&\cite{xu2015classifying}&+ POS + NER + WordNet&55.3\\

\cline{2-4}

&DepNN&Word embeddings, WordNet&55.2\\
&\cite{liu2015dependency}&&\\

\cline{2-4}

&BRCNN&Word embeddings&55.0\\
&\cite{cai2016bidirectional}&+ POS, NER, WordNet&55.6\\

\hline

\hline
\multicolumn{1}{c|}{\multirow{2}{*}{\textbf{Our Model}}}

&SR-BRCNN&Word embeddings&65.2\\
&&+ POS, NER, WordNet&65.9\\

\cline{2-4}

\hline

\hline
\end{tabular}

\caption{Comparison of relation classification systems on the dataset of Sanwen}\label{daimprovements}
\end{table*}

\subsection{Various Structure Regularization Methods}

As mentioned in the former subsection, a certain rule is needed for selecting nodes from the dependency tree to cut the subtrees. Here we have experimented several ways to determine whether a node should be selected.

The punctuation is a natural break point of the sentence. The most intuitive method to cut the dependency trees is to cut by punctuation. And the resulting subtrees usually keep similar syntax to traditional dependency trees. One another popular method to regularize the structure is to decompose the structure randomly. In our model, we will randomly select several nodes in the dependency tree and then cut the subtrees under these nodes. Finally we decide to cut the dependency tree by prepositions, especially in Chinese Sanwen. There usually are many decorations to describe the entities, and the using of prepositional phrases is very common for that purpose. So we also tried decomposing the dependency trees using prepositions.

\section{Experiments}
We evaluated our model on two datasets. One is the Chinese Sanwen dataset, which is the main platform we conduct our experiment. The other one is the SemEval2010 Task 8 dataset, which is an established benchmark for relation classification. The dataset contains 8000 sentences for training, and 2717 for testing. We split 800 samples out of the training set for validation.

\subsection{Dataset}
We evaluate our model on two datasets, the Sanwen dataset and the SemEval2010 Task 8 dataset. 

The Sanwen dataset has (K+1)=10 distinguished relations, as follows.
\begin{itemize}
	\item Located, Near, Part-Whole, Family, Social, Create, Use, Ownership, General-Special
	\item Null
\end{itemize}

The SemEval2010 Task 8 dataset, which is an established benchmark for relation classification \cite{Hendrickx2010}. The dataset contains 8000 sentences for training, and 2717 for testing. We split 800 samples out of the training set for validation. 

The dataset also has (K+1)=10 distinguished relations, as follows.
\begin{itemize}
	\item Cause-Effect, Component-Whole, Content-Container, Entity-Destination, Entity-Origin, Message-Topic, Member-Collection, Instrument-Agency, Product-Agency
	\item Other
\end{itemize}

The former K=9 relations are directed, with the Other Class undirected. There are (2K+1)=19 different classes for 10 relations. All baseline systems and our model use the official macro-averaged F1-score to evaluate model performance. This official measurement excludes the Other/Null relation.

\subsection{Experiment settings}
In our experiment, we take mostly the same parameters as Cai et al. \shortcite{cai2016bidirectional}. We use pre-trained word embeddings, which are trained on Gigaword with word2vec(\cite{mikolov2013distributed}). Word embeddings are 200-dimensional. The embeddings of relation are initialized randomly and are 50-dimensional. The hidden layer of LSTMs to extract information from entities and relations are the same as the embedding dimension of entities and relations. According to the work of Cai et al. \shortcite{cai2016bidirectional}, the performance of BRCNN is improved when we assign different embeddings to relation and the reverse of the relation. Thus we keep the settings.

We applied l2 regularization to weights in neural networks and dropout to embeddings with a keep probability 0.5. AdaDelta \cite{zeiler2012adadelta} is used for optimization.

\subsection{Experimental Results}
Table 1 compares our SR-BRCNN model with other state-of-the-art methods on the corpus of Sanwen. The first entry in the table presents the highest performance achieved by traditional feature-based methods. Hendrick et al. \shortcite{Hendrickx2010} fed a variety of handcrafted features to the SVM classifier and achieve an $F_{1}$-score of 48.9.

Recent performance improvements on the task of relation classification are mostly achieved with the help of neural networks. Socher et al. \shortcite{socher2011semi} built a recursive neural network on the constituency tree and achieved a comparable performance with Hendrick et al. \shortcite{Hendrickx2010}.  Xu et al. \shortcite{xu2015classifying} introduced a type of gated recurrent neural network (LSTM) which could raise the $F_{1}$ score to 55.3.

From the perspective of convolution, Zeng et al. \shortcite{zeng2014relation} constructed a CNN on the word sequence; they also integrated word position embeddings, which helped a lot on the CNN architecture. dos Santos et al. \shortcite{santos2015classifying} proposed a similar CNN model, named CR-CNN, by replacing the common softmax cost function with a ranking-based cost function. By diminishing the impact of the Other class, they have achieved an $F_{1}$-score of 54.1. Along the line of CNNs, Liu et al. \shortcite{liu2015dependency} proposed a convolutional neural network with a recursive neural network designed to model the subtrees, and achieve an $F_{1}$-score of 55.2.

\begin{table}[h]
		\centering
		\begin{tabular}{|l|c|}\hline
			Classifier&$F_{1}$ score\\\hline
			BRCNN&55.6\\
			SR-BRCNN&65.9\\
			\hline
		\end{tabular}
		\caption{Results on Sanwen}
		\label{tab:Margin_settings}
	\end{table}

\begin{table}[h]
		\centering
		\begin{tabular}{|l|c|}\hline
			Classifier&$F_{1}$ score\\\hline
			BRCNN&84.6\\
			SR-BRCNN&85.1\\
			\hline
		\end{tabular}
		\caption{Results on SemEval2010 Task 8}
		\label{tab:Margin_settings}
	\end{table}

Table 2 and Table 3 compare our model with the basic BRCNN, Cai et al.\shortcite{cai2016bidirectional} methods on the corpus of Sanwen and SemEval2010 Task 8 dataset. On both corpus, structure regularization helps improve the result, especially on the corpus of Sanwen. The basic BRCNN uses the SDP extracted from the dependency tree, regardless of the quality of it. The method of structure regularization could prevent the overfitting of SDPs with low quality.

\begin{table}[h]
		\centering
		\begin{tabular}{|l|c|}\hline
			Classifier&$F_{1}$ score\\\hline
			BRCNN&84.6\\
			SR by punctuation &84.5\\
			SR by random &84.7\\
			SR by preposition &85.1\\
			\hline
		\end{tabular}
		\caption{Different SR Results on SemEval}
		\label{tab:Margin_settings}
	\end{table}

\begin{table}[h]
		\centering
		\begin{tabular}{|l|c|}\hline
			Classifier&$F_{1}$ score\\\hline
			BRCNN&55.6\\
			SR by punctuation &59.7\\
			SR by random &62.4\\
			SR by preposition &65.9\\
			\hline
		\end{tabular}
		\caption{Different SR Results on Sanwen}
		\label{tab:Margin_settings}
	\end{table}

\subsection{Analysis: effect of SR:}

\begin{figure}
	\centering
	\includegraphics[width= 6.5cm]{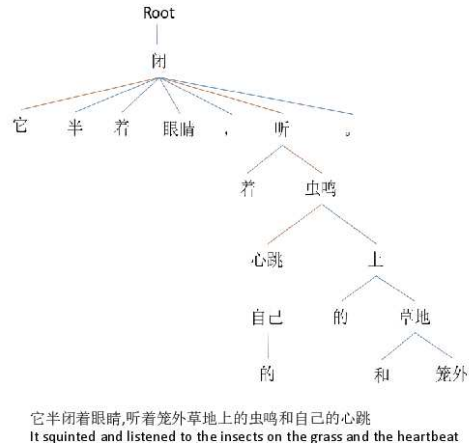}
	\caption{The SDP of the origin dependency tree}\label{fig4}
\end{figure}

\begin{figure}
	\centering
	\includegraphics[width= 8cm]{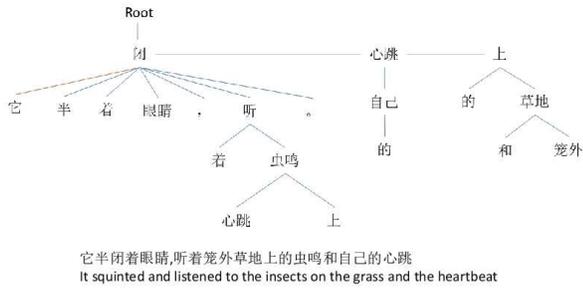}
	\caption{The SDP of the structure regularized dependency tree}\label{fig5}
\end{figure}

Figure~\ref{fig4} and Figure~\ref{fig5} show an example of structure regularized SDP. The red lines show the SDP in the origin dependency tree and the SR-SDP in the structure regularized tree. The main idea of the method is to avoid the incorrect structure from the dependency trees generated by the parser. The SDP in Figure~\ref{fig4} is longer than the SR-SDP in Figure~\ref{fig5}. However, the dependency tree of the example is not completely correct. The longer the SDP is, the more incorrect information the model learns.

The structure regularized BRCNN has shown obvious improvements on both English and Chinese datasets. A question is why structure regularization works much better than original models. We attribute the improvements to the simplified structures that generated by structure regularization. As we discussed before, long distance of two entities will weaken the suggestions of their relations. And the internal relations of components of a sentence are more obscure due to the Sanwen’s natural attribute and characteristic. Thus explicit structures is necessary. By conducting structure regularization on the dependency tree, we get several subtrees with more simpler structure then we extract SDP from the lined forests. Apparently, the distance between two entities has a great chance to be shortened. In most cases, the distance between two entities will be shortened along the new SR-SDP. Without the redundant information along the original SDP. The BRCNN that benefits from the intensive dependencies will capture more effective information for classification.

\subsection{Analysis: effect of different regularization methods.}
The punctuation is a natural break point of the sentence. The advantage of decomposing the dependency tree by punctuation is that subtrees are more like the traditional dependency trees in the aspect of integrity. However, not every sentence has punctuations. Even for the sentences with punctuations, the original dependency trees can not be sufficiently regularized. Despite its drawbacks, cutting the dependency trees is by punctuation shows obvious improvements on the model and leads to further experiments.

Another method is inherited from the idea of structure regularization. We just regularize the structure by decomposing the structure randomly. This method will solve the insufficient problems mentioned above. Superficially, this approach may cause the loss of the information provided by the structure of the dependency trees. Actually, method of structure regularization has shown that the this degree of loss of information is not a serious problem. As is shown in our experiments, the random method regularize the structure sufficiently and it gives a slightly better result compared to cutting dependency trees by punctuation.

A more elaborate method is to cut the dependency tree by prepositions. In Chinese Sanwen, there usually are many decoration to describe the entities, so prepositional phrases are used frequently, even more than punctuations. Cutting by prepositions has two advantages. First, compared to cutting the dependency trees by punctuation, it will regularize the tree more sufficiently. Second, compared to cut the dependency tree randomly, the subtrees under the prepositional nodes are usually internally linked. The decomposition based on preposition overcomes the drawback of random regularization to a certain degree.

\section{Related Work}

Relation classification plays an important role in NLP. Traditional methods are usually feature-based. and their performance strongly depends on the quality of the extracted features. Kambhatla et al. \shortcite{Kambhatla2004} used a maximum entropy model for feature combination. Hendrick et al. \shortcite{Hendrickx2010} collected various features, including lexical, syntactic as well as semantic features. 

In kernel based methods, similarity between two data samples is measured without explicit feature
representation. Bunescu and Mooney et al. \shortcite{bunescu-mooney2005} designed a kernel along the shortest dependency path between two entities by observing that the relation strongly relies on SDPs.  Wang et al. \shortcite{wang2008re} provide an analysis of the relative strength and weakness of several kernels through systematic and showed that relation extraction can benefit from combining convolution kernel and syntactic features.  Plank and Moschitti et al\shortcite{plank2013embedding} propose to combine (i) term generalization approaches such as word clustering and latent semantic analysis (LSA) and (ii) structured kernels to improve the adaptability of relation extractors to new text genres/domains. 

Yu et al. \shortcite{yu2014factor} capitalizes on arbitrary types of linguistic annotations by better utilizing features associated with substructures of those annotations, including global information.

Recently, deep neural networks are widely used in relation classification. works are widely used in relation classification. Zeng et al. \shortcite{zeng2014relation} exploit a convolutional deep neural network to extract lexical and sentence level features.  Wang et al. \shortcite{wang2016relation} proposes a convolutional neural network with two lever of attentions. The attention mechanism helps capture both entity-specific attention and relation-specific attention, thus more subtle cues can be detected.

Not only convolutional networks, recurrent nerworks have been proposed to address this. Zhang et al. \shortcite{zhang2015bidirectional} In this work, bidirectional long short-term memory networks are used to model the sentence with sequential information. Miwa et al. \shortcite{miwaend} present an end-to-end neural model to extract entities and relations between them. Both word sequence and dependency tree information can be captured by stacking tree-structured LSTM-RNNs on sequential LSTM-RNNs. Xu et al.\shortcite{xusemantic} used a CNN to learn the relation from SDP. Nguyen et al. \shortcite{nguyen2015combining} combine the traditional feature-based method, the convolutional and recurrent neural networks to benefit from their advantages. Cai et al. \shortcite{cai2016bidirectional} propose BRCNN to model the SDP, which can pick up bidirectional information with a combination of LSTM and CNN. There are methods for optimizing the training process, such as 
\citeauthor{SunLWL14} (\citeyear{SunLWL14}) and \citeauthor{Sun2016Asynchronous} (\citeyear{Sun2016Asynchronous}).

\section{Conclusions}

In this paper, we present a novel model, Structure Regularized Bidirectional Recurrent Convolutional Neural Network, to classify the relation of two entities in a sentence. We prove that tree-based structure regularization can help improve the results, while the method are normally used in sequence-based models before. The results also show that how different ways of regularization act in the model of BRCNN. The best way of them helps improve the $F_{1}$ score by 10.3.
We also develop a corpus on Chinese Sanwen focusing on the task of Named Entity Recognition and Relation Classification. Researchers often find it difficult to conduct their experiments on Chinese, due to the lack of Chinese corpus in different fields. The Sanwen corpus is large enough for us to train models and verify the models.

\bibliography{ref}
\bibliographystyle{aaai}
\end{CJK*}
\end{document}